%% file: iclr2017_conference.tex
\documentclass{article} 
\usepackage{iclr2017_conference,times}
\usepackage{hyperref}
\usepackage{url}

\usepackage{enumitem}

\usepackage[utf8]{inputenc}

\usepackage{algorithm}
\usepackage{algpseudocode}
\usepackage{graphicx}
\usepackage{subfig}
\usepackage{hyperref}
\usepackage[mmddyy,hhmmss]{datetime}

\usepackage{amstext,amssymb,amsmath}
\usepackage{amsthm}
\newtheorem{thm}{Theorem}
\newtheorem{lem}[thm]{Lemma}
\newtheorem{definition}{Definition}

\newcommand{\boldpara}[1] {\vspace*{0.075in}\noindent\textbf{#1:}}

\title{Semi-supervised Knowledge Transfer \\ for Deep Learning from Private Training Data}

\author{Nicolas Papernot\thanks{Work done while the author was at Google.} \\
Pennsylvania State University\\
\texttt{ngp5056@cse.psu.edu} 
\And
Martín Abadi \\
Google Brain \\
\texttt{abadi@google.com} 
\And
Úlfar Erlingsson \\
Google \\
\texttt{ulfar@google.com} 
\And
Ian Goodfellow \\
Google Brain\thanks{Work done both at Google Brain and at OpenAI.}\\
\texttt{goodfellow@google.com}
\And
Kunal Talwar \\
Google Brain\\
\texttt{kunal@google.com}
\And
\hspace*{1.15in} \\
~ \\
~
}

\iclrfinalcopy 

\begin{document}

\maketitle

\begin{abstract}
\input{abstract}
\end{abstract}

\input{introduction}

\input{approach}
\input{analysis-merged}
\input{evaluation}

\input{related-work}
\input{conclusions}

\subsubsection*{Acknowledgments}

Nicolas Papernot is supported by a Google PhD Fellowship in Security. The
authors would like to thank Ilya Mironov and Li Zhang for insightful discussions
about early drafts of this document.


\newpage

\appendix
\newpage
\input{ap-privacy-analysis-short}
\newpage
\input{ap-learning-student}
\newpage
\input{ap-uci}

\end{document}

%% file: abstract.tex
Some machine learning applications involve training data that is sensitive, such
as the medical histories of patients in a clinical trial. A model may
inadvertently and implicitly store some of its training data; careful analysis
of the model may therefore reveal sensitive information.

To address this problem, we demonstrate a generally applicable approach to
providing strong privacy guarantees for training data: \emph{Private Aggregation of Teacher Ensembles} (PATE). The approach combines, in
a black-box fashion, multiple models trained with disjoint datasets, such as
records from different subsets of users. Because they rely directly on sensitive
data, these models are not published, but instead used as ``teachers'' for a
``student'' model. The student learns to predict an output chosen by noisy voting
among all of the teachers, and cannot directly access an individual teacher or
the underlying data or parameters. The student's privacy properties can be
understood both intuitively (since no single teacher and thus no single dataset
dictates the student's training) and formally, in terms of differential privacy.
 These properties hold even if an adversary can not only query the student but
also inspect its internal workings.

Compared with previous work, the approach imposes only weak assumptions on how
teachers are trained: it applies to any model, including non-convex models like
DNNs. We achieve state-of-the-art privacy/utility trade-offs on MNIST and SVHN
thanks to an improved privacy analysis and semi-supervised learning.

%% file: introduction.tex
\section{Introduction} \label{sec:introduction}


Some machine learning applications with great benefits are enabled only through
the analysis of sensitive data, such as users' personal contacts, private
photographs or correspondence, or even medical records or genetic
sequences~\citep{alipanahi2015predicting,kannan2016smart,kononenko2001machine,sweeney1997weaving}. Ideally, in those cases, the learning algorithms would protect the privacy of users' training data, e.g., by guaranteeing that the output model generalizes away from the specifics of any individual user. %
Unfortunately, established machine learning algorithms make no such guarantee;
indeed, though state-of-the-art algorithms generalize well to the test set, they
continue to overfit on specific training examples in the sense that some of
these examples are implicitly memorized.

Recent attacks exploiting this implicit memorization in machine learning have
demonstrated that private, sensitive training data can be recovered from models.
Such attacks can proceed directly, by analyzing internal model parameters, but
also indirectly, by repeatedly querying opaque models to gather data for the
attack's analysis. For example, \citet{fredrikson2015model} used hill-climbing
on the output probabilities of a computer-vision classifier to reveal individual
faces from the training data. Because of those demonstrations---and because
privacy guarantees must apply to worst-case outliers, not only the average---any
strategy for protecting the privacy of training data should prudently assume
that attackers have unfettered access to internal model parameters.

To protect the privacy of training data, this paper improves upon a specific,
structured application of the techniques of knowledge aggregation and
transfer~\citep{ML:Breiman:bagging}, previously explored
by~\citet{nissim2007smooth},~\citet{pathak2010multiparty}, and
particularly~\citet{hamm2016learning}. In this strategy, first, an
ensemble~\citep{dietterich2000ensemble} of teacher models is trained on disjoint subsets of the sensitive data. Then,
using auxiliary, unlabeled non-sensitive data, a student model is trained on the
aggregate output of the ensemble, such that the student learns to accurately
mimic the ensemble. Intuitively, this strategy ensures that the student does not
depend on the details of any single sensitive training data point (e.g., of any
single user), and, thereby, the privacy of the training data is protected even
if attackers can observe the student's internal model parameters.

This paper shows how this strategy's privacy guarantees can be strengthened by
restricting student training to a limited number of teacher votes, and by
revealing only the topmost vote after carefully adding random noise. %
We call this strengthened strategy PATE, for \emph{Private Aggregation of Teacher Ensembles}.
Furthermore, we introduce an improved privacy analysis that makes the strategy
generally applicable to machine learning algorithms with high utility and
meaningful privacy guarantees---in particular, when combined with
semi-supervised learning.

To establish strong privacy guarantees, it is important to limit the student's
access to its teachers, so that the student's exposure to teachers' knowledge
can be meaningfully quantified and bounded. Fortunately, there are many
techniques for speeding up knowledge transfer that can reduce the rate of
student/teacher consultation during learning. We describe several techniques
in this paper, the most effective of which makes use of generative adversarial
networks (GANs)~\citep{goodfellow2014generative} applied to semi-supervised learning,
using the implementation proposed
by~\citet{salimans2016improved}.
For clarity,
we use the term PATE-G
when our approach is combined with generative, semi-supervised methods.
Like all semi-supervised learning
methods, PATE-G assumes the student has access to additional, unlabeled
data, which, in this context, must be public or non-sensitive. This assumption
should not greatly restrict our method's applicability: even when learning on
sensitive data, a non-overlapping, unlabeled set of data often exists, from
which semi-supervised methods can extract distribution priors. For instance,
public datasets exist for text and images, and for medical
data.

It seems intuitive, or even obvious, that a student machine learning model will
provide good privacy when trained without access to sensitive training data,
apart from a few, noisy votes from a teacher quorum. However, intuition is not
sufficient because privacy properties can be surprisingly hard to reason about;
for example, even a single data item can greatly impact machine learning models
trained on a large corpus~\citep{chaudhuri2011differentially}. Therefore, to
limit the effect of any single sensitive data item on the student's learning,
precisely and formally, we apply the well-established, rigorous standard of
differential privacy~\citep{dwork2014algorithmic}. Like all differentially
private algorithms, our learning strategy carefully adds noise, so that the
privacy impact of each data item can be analyzed and bounded. In particular, we
dynamically analyze the sensitivity of the teachers' noisy votes; for this
purpose, we use the state-of-the-art moments accountant technique
from~\citet{abadi2016deep}, which tightens the privacy bound when the topmost
vote has a large quorum. As a result, for MNIST and similar benchmark learning
tasks, our methods allow students to provide excellent utility, while our
analysis provides meaningful worst-case guarantees. In particular, we can bound
the metric for privacy loss (the differential-privacy $\varepsilon$) to a range
similar to that of existing, real-world privacy-protection mechanisms, such as
Google's RAPPOR~\citep{erlingsson2014rappor}.

Finally, it is an important advantage that our learning strategy and our privacy
analysis do not depend on the details of the machine learning techniques used to
train either the teachers or their student. Therefore, the techniques in this
paper apply equally well for deep learning methods, or any such learning methods
with large numbers of parameters, as they do for shallow, simple techniques. In
comparison, \citet{hamm2016learning} guarantee privacy only conditionally, for a
restricted class of student classifiers---in effect, limiting applicability to
logistic regression with convex loss. Also, unlike the methods
of~\citet{abadi2016deep}, which represent the state-of-the-art in
differentially-private deep learning, our techniques make no assumptions about
details such as batch selection, the loss function, or the choice of the optimization algorithm. Even so, as we show in experiments on MNIST
and SVHN, our techniques provide a privacy/utility tradeoff that equals or
improves upon bespoke learning methods such as those of~\citet{abadi2016deep}.

Section~\ref{sec:related-work} further discusses the related work. Building on
this related work, our contributions are as follows:
\begin{itemize}[itemsep=1pt, topsep=0pt, partopsep=0pt]
  
\item We demonstrate a general machine learning strategy, the PATE approach, that provides
  differential privacy for training data in a ``black-box'' manner, i.e., independent of the learning algorithm, as demonstrated by Section~\ref{sec:evaluation} and Appendix~\ref{ap:uci}.
	
\item We improve upon the strategy outlined in~\citet{hamm2016learning}
  for learning machine models that protect training data privacy.
  In particular, 
  our student only accesses the teachers' top vote and the model does not
  need to be trained with a restricted class of convex losses.

\item We explore four different approaches for 
  reducing the student's dependence on its teachers, and show how the
   application of GANs
  to semi-supervised learning of~\citet{salimans2016improved} can greatly reduce the privacy loss by radically reducing the need for supervision. 

\item We present a new application of the 
  moments accountant technique from \citet{abadi2016deep}
  for improving the differential-privacy analysis of knowledge transfer,
  which allows the training of students with meaningful privacy bounds.

\item We evaluate our framework on MNIST and SVHN, allowing for a comparison of
	our results with previous differentially private machine learning methods. Our
	classifiers achieve an $(\varepsilon,\delta)$ differential-privacy bound of
	$(2.04,10^{-5})$ for MNIST and $(8.19,10^{-6})$ for SVHN, respectively with
	accuracy of $98.00\%$ and $90.66\%$. In comparison, for
	MNIST,~\citet{,abadi2016deep} obtain a looser $(8, 10^{-5})$ privacy bound and
	$97\%$ accuracy. For SVHN,~\citet{shokri2015privacy} report approx.\ $92\%$
	accuracy with $\varepsilon>2$ per each of $\mathrm{300\mbox{,}000}$ model
	parameters, naively making the total $\varepsilon>\mathrm{600\mbox{,}000}$,
	which guarantees no meaningful privacy.

\item Finally, we show that the PATE approach can be successfully applied
        to other model structures and to datasets with different characteristics.
      In particular, in Appendix~\ref{ap:uci}
      PATE protects the privacy of medical data
      used to train a model based on random forests.
\end{itemize}

Our results are encouraging, and highlight the benefits of combining a learning
strategy based on semi-supervised knowledge transfer with a precise,
data-dependent privacy analysis. However, the most appealing aspect of this work
is probably that its guarantees can be compelling to both an expert and a
non-expert audience. %
In combination, our techniques simultaneously provide both an intuitive and a
rigorous guarantee of training data privacy, without sacrificing the utility of
the targeted model. This gives hope that users will increasingly be able to
confidently and safely benefit from machine learning models built from their
sensitive data.

%% file: approach.tex
\section{Private learning with ensembles of teachers}
\label{sec:approach}

\begin{figure}[t]
  \centering
  \includegraphics[width=\columnwidth]{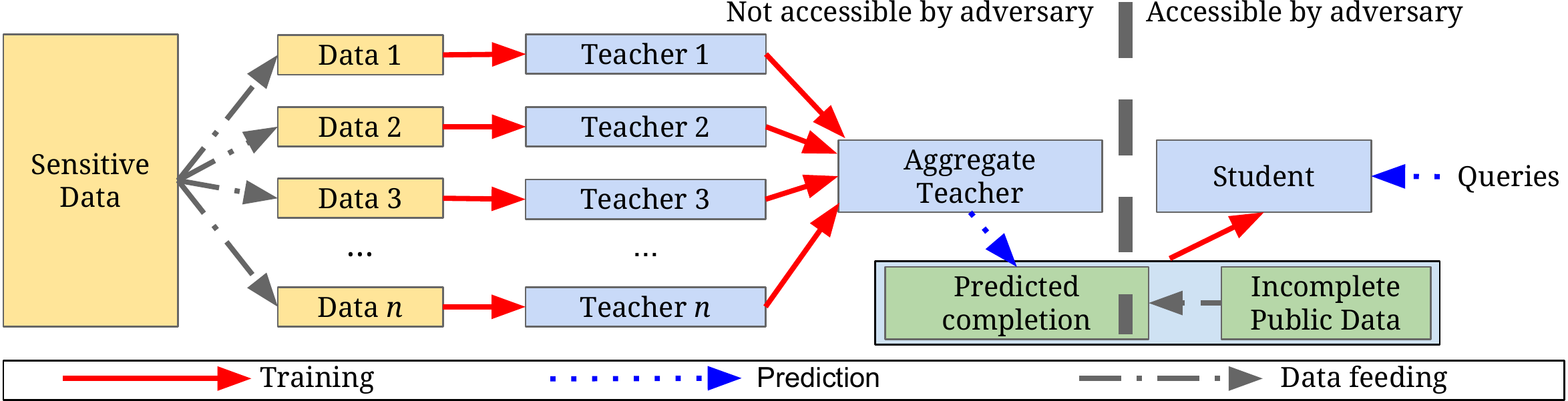}
  \caption{Overview of the approach: (1) an ensemble of teachers 
  is trained on disjoint subsets of the sensitive data, (2) a 
  student model is trained on public data labeled using the ensemble.}
  \label{fig:approach-overview}
\end{figure}

In this section, we introduce the specifics of the PATE approach, which is illustrated in Figure~\ref{fig:approach-overview}. We describe how the data is
partitioned to train an ensemble of teachers, and how the predictions made by
this ensemble are noisily aggregated. In addition, we discuss how GANs can be used in training the student, and distinguish PATE-G variants that improve our approach using generative, semi-supervised methods.

\subsection{Training the ensemble of teachers} 
\label{ssec:teacher}

\boldpara{Data partitioning and teachers} Instead of training a single model to
solve the task associated with dataset $(X,Y)$, where $X$ denotes the set of
inputs, and $Y$ the set of labels, we partition the data in $n$ disjoint sets
$(X_n, Y_n)$ and train a model separately on each set. As evaluated in
Section~\ref{ssec:eval-teacher}, assuming that $n$ is not too large with respect
to the dataset size and task complexity, we obtain $n$ classifiers $f_i$ called
teachers. We then deploy them as an ensemble making predictions on unseen inputs
$x$ by querying each teacher for a prediction $f_i(x)$ and aggregating these
into a single prediction.

\boldpara{Aggregation} The privacy guarantees of this teacher ensemble stems
from its aggregation. Let $m$ be the number of classes in our task. The label
count for a given class $j\in [m]$ and an input $\vec{x}$ is the number of
teachers that assigned class $j$ to input $\vec{x}$: $n_j(\vec{x}) = \left| \{
i: i\in [n], f_i(\vec{x}) = j \} \right|$. If we simply apply
\emph{plurality}---use the label with the largest count---the ensemble's
decision may depend on a single teacher's vote. Indeed, when two labels have a
vote count differing by at most one, there is a tie: the aggregated output
changes if one teacher makes a different prediction. We add random
noise to the vote counts $n_j$ to introduce ambiguity:
\begin{equation}
\label{eq:noisy-max} f(x) = \arg\max_j \left\{n_j(\vec{x}) +
Lap\left(\frac{1}{\gamma}\right)\right\}
\end{equation}
In this equation, $\gamma$ is a privacy parameter and $Lap(b)$ the
Laplacian distribution with location $0$ and scale $b$. The parameter
  $\gamma$ influences the privacy guarantee we can prove. Intuitively,
  a large $\gamma$ leads to a strong privacy guarantee, but can
  degrade the accuracy of the labels, as the noisy maximum $f$ above
  can differ from the true plurality.

While we could use an $f$ such as above to make
  predictions, the noise required would increase as we make more
  predictions, making the model useless after a bounded number of queries. Furthermore, privacy guarantees do not hold when an adversary has
access to the model parameters. Indeed, as each teacher $f_i$ was trained without taking
into account privacy, it is conceivable that they have sufficient capacity to
retain details of the training data. To address these limitations, we train
another model, the student, using a fixed number of labels predicted by the
teacher ensemble.

\subsection{Semi-supervised transfer of the knowledge from an ensemble to a student}

We train a student on nonsensitive and unlabeled data, some of which we label
using the aggregation mechanism. This student model is the one deployed, in lieu
of the teacher ensemble, so as to fix the privacy loss to a value that does not grow with the number of user queries made to the student model.
Indeed, the privacy loss is now determined by the number of queries made to the
teacher ensemble during student training and does not increase as end-users
query the deployed student model. Thus, the privacy of users who contributed to
the original training dataset is preserved even if the student's architecture
and parameters are public or reverse-engineered by an adversary.

We considered several techniques to trade-off the student model's quality with the number of labels it needs to access: distillation, active
learning,  semi-supervised learning (see Appendix~\ref{ap:student-learning}).
Here, we only describe the most successful one, used in PATE-G: semi-supervised learning with GANs.

\boldpara{Training the student with GANs} The GAN framework involves two machine
learning models, a \emph{generator} and a \emph{discriminator}. They are trained
in a competing fashion, in what can be viewed as a two-player
game~\citep{goodfellow2014generative}. The generator produces samples from the
data distribution by transforming vectors sampled from a Gaussian distribution.
The discriminator is trained to distinguish samples artificially produced by the
generator from samples part of the real data distribution. Models are trained
via simultaneous gradient descent steps on both players' costs. In practice, these dynamics are often difficult to control
when the strategy set is non-convex (e.g., a DNN). 
In their application of
GANs to semi-supervised learning, \citet{salimans2016improved} made
 the following modifications. The
discriminator is extended from a binary classifier (data vs.~generator sample)
to a multi-class classifier (one of $k$ classes of data samples, plus a class
for generated samples). This classifier is then trained to classify 
labeled real samples
in the correct class, unlabeled real samples in any of the $k$ classes, and the generated samples in the additional
class.

Although no formal results currently explain why yet, the technique was
empirically demonstrated to greatly improve semi-supervised learning of
classifiers on several datasets, especially when the classifier is trained with
{\em feature matching} loss~\citep{salimans2016improved}.

Training the student in a semi-supervised fashion makes better use of the entire
data available to the student, while still only labeling a subset of it.
Unlabeled inputs are used in unsupervised learning to estimate a good prior for
the distribution. Labeled inputs are then used for supervised
learning. 

%% file: analysis-merged.tex
\section{Privacy analysis of the approach}
\label{sec:dp-analysis}
\newcommand{\ifnoisymax}[2]{#1}
\newcommand{\eps}{\varepsilon}
\newcommand{\M}{\mathcal{M}}
\newcommand{\Domain}{\mathcal{D}}
\newcommand{\Range}{\mathcal{R}}
\newcommand{\outcome}{o}
\newcommand{\eqdef}{\stackrel{\Delta}=}
\newcommand{\E}{\mathbb{E}}
\newcommand{\comment}[1]{}

We now analyze the differential privacy guarantees of our PATE approach.
Namely, we keep track of the privacy budget throughout the student's
training using the moments accountant~\citep{abadi2016deep}. When teachers reach
a strong quorum, this allows us to bound privacy costs more strictly.

\subsection{Differential Privacy Preliminaries and  a Simple Analysis of \mbox{PATE}}
Differential privacy~\citep{dwork2006calibrating,dwork2011firm} has established
itself as a strong standard. It provides privacy guarantees for algorithms
analyzing databases, which in our case is a machine learning training algorithm
processing a training dataset. Differential privacy is defined using pairs of
adjacent databases: in the present work, these are datasets that only differ by
one training example. Recall the following variant of differential privacy
introduced in~\cite{ODO}.
\begin{definition} 
A randomized mechanism $\M$ with domain $\Domain$ and range
$\mathcal{R}$ satisfies $(\eps,\delta)$-differential privacy if for
any two adjacent inputs $d, d'\in \Domain$ and for any subset of
outputs $S\subseteq\Range$ it holds that: 
	\begin{equation}
	\label{eq:dp}
	\Pr[\M(d)\in S]\leq e^{\eps}\Pr[\M(d')\in S]+\delta.
	\end{equation}
\end{definition}

It will be useful to define the \emph{privacy loss} and the \emph{privacy loss
random variable}. They capture the differences in the probability distribution
resulting from running $\M$ on $d$ and $d'$.

\begin{definition}
Let $\M \colon \Domain \rightarrow \Range$ be a randomized mechanism and $d, d'$
a pair of adjacent databases. Let \textsf{aux} denote an auxiliary input. For an
outcome $\outcome \in \Range$, the privacy loss at $\outcome$ is defined as:
\begin{equation}
  c(\outcome; \M,  \textsf{aux}, d, d') \eqdef \log \frac{\Pr[\M( \textsf{aux}, d) = \outcome]}{\Pr[\M( \textsf{aux}, d') = \outcome]}.
\end{equation}
The privacy loss random variable $C(\M, \textsf{aux}, d, d')$ is
defined as $c(\M(d); \M, \textsf{aux}, d, d')$, i.e. the random
variable defined by evaluating the privacy loss at an outcome sampled
from $\M(d)$.
\end{definition}

A natural way to bound our approach's privacy loss is to first bound the
privacy cost of each label queried by the student, and then use the strong
composition theorem~\citep{DRV10} to derive the total cost of training the
student. For neighboring databases $d, d'$, each teacher gets the same
training data partition (that is, the same for the teacher with $d$ and
with $d'$, not the same across teachers), with the exception of one teacher
whose corresponding training data partition differs. 
Therefore, the label counts $n_j(\vec{x})$ for any example $\vec{x}$,
on $d$ and $d'$ differ by at most $1$ in at most two locations. In the
next subsection, we show that this yields loose guarantees.

\subsection{The moments accountant: A building block for better analysis}

To better keep track of the privacy cost, we use recent advances in privacy cost
accounting. The moments accountant was introduced by~\cite{abadi2016deep},
building on previous work~\citep{BunS16, DworkR16, Mironov16}. 

\begin{definition}
	Let $\M \colon \Domain \rightarrow \Range$ be a randomized mechanism and $d, d'$ a pair of adjacent databases. Let \textsf{aux} denote an auxiliary input. The moments accountant is defined as:
	\begin{equation}\label{eq:moments-accountant}
	\alpha_\M(\lambda) \eqdef \max_{\textsf{aux}, d, d'} \alpha_\M(\lambda;\textsf{aux},d,d')
	\end{equation}
	where $\alpha_\M(\lambda;\textsf{aux},d,d') \eqdef 
	\log \E[\exp(\lambda C(\M, \textsf{aux}, d, d'))]$ is the
        moment generating function of the privacy loss random variable.
\end{definition}

The following properties of the moments accountant are proved 
in~\cite{abadi2016deep}.
\begin{thm}\label{thm:property}
	1. \textbf{[Composability]}
	Suppose that a mechanism $\M$ consists of a sequence of adaptive mechanisms $\M_1, \ldots, \M_k$ where $\M_i\colon \prod_{j=1}^{i-1}\Range_j\times \Domain \to\Range_i$. Then, for any output sequence $\outcome_1,\dots,\outcome_{k-1}$ and any $\lambda$
	\[
	\alpha_\M(\lambda;d,d') = \sum_{i=1}^k \alpha_{\M_i}(\lambda;\outcome_1,\dots,\outcome_{i-1},d,d')\,,\]
	where $\alpha_\M$ is conditioned on $\M_i$'s output being $\outcome_i$ for $i<k$. 
	
	2. \textbf{[Tail bound]}
	For any $\eps>0$, the mechanism $\M$ is $(\eps, \delta)$-differentially private for
	\[\delta=\min_{\lambda} \exp(\alpha_\M(\lambda) -\lambda \eps)\,.\]
\end{thm}

We write down two important properties of the aggregation mechanism from Section~\ref{sec:approach}. The 
first property is proved in~\cite{dwork2014algorithmic}, and the second follows 
from~\cite{BunS16}.

\begin{thm}
  \label{thm:noisymax}
  Suppose that on neighboring databases $d, d'$, the label counts $n_j$ differ by at most 1 in each coordinate. Let $\M$ be the mechanism that reports $\arg\max_{j} \left\{n_j + Lap(\frac 1 \gamma) \right\}$. Then $\M$ satisfies $(2\gamma, 0)$-differential privacy. Moreover, for any $l$, \textsf{aux}, $d$ and $d'$,
  \begin{align}
    \alpha(l; \textsf{aux}, d, d') \leq 2\gamma^2 l(l+1) \label{eqn:lmgf_basic}
  \end{align}
\end{thm}

At each step, we use the \ifnoisymax{aggregation mechanism with noise $Lap(\frac
1 \gamma)$}, which is $(2\gamma,0)$-DP. Thus over $T$ steps, we get $(4T\gamma^2
+ 2\gamma\sqrt{2T\ln \frac 1 \delta}, \delta)$-differential privacy. This can be
rather large: plugging in values that correspond to our SVHN result,
$\gamma=0.05, T=1000, \delta= 1\mathrm{e}{-6}$ gives us $\eps \approx 26$ or
alternatively plugging in values that correspond to our MNIST result,
$\gamma=0.05, T=100, \delta= 1\mathrm{e}{-5}$ gives us $\eps \approx 5.80$.

\subsection{A precise, data-dependent privacy analysis  of \mbox{PATE}}
Our data-dependent privacy analysis takes advantage of the fact that when the quorum
among the teachers is very strong, the majority outcome has overwhelming
likelihood, in which case the privacy cost is small whenever this outcome
occurs. The moments accountant allows us analyze the composition of such
mechanisms in a unified framework.

The following theorem, proved in Appendix~\ref{ap:privacy-analysis}, provides a data-dependent bound on the moments
of any differentially private mechanism where some specific outcome is
very likely.

\begin{thm}
  \label{thm:lmgf_data_dep}
	Let $\M$ be $(2\gamma, 0)$-differentially private and $q \geq \Pr[\M(d) \neq \outcome^*]$ for some outcome $\outcome^*$. Let $l,\gamma \geq 0$ and $q <\frac{e^{2\gamma}-1}{e^{4\gamma}-1}$. Then for any $\textsf{aux}$ and any neighbor $d'$ of $d$, $\M$ satisfies
	\begin{align*}
	\alpha(l; \textsf{aux}, d, d') \leq \log ((1-q)\Big(\frac{1-q}{1-e^{2\gamma}q}\Big)^l + q\exp(2\gamma l)).
	\end{align*}
\end{thm}

To upper bound $q$ for our aggregation mechanism, we use the following simple lemma, also proved in Appendix~\ref{ap:privacy-analysis}.
\begin{lem}
 \label{lem:bound_on_q}
  Let $\mathbf{n}$ be the label score vector for a database $d$ with $n_{j^*} \geq n_j$ for all $j$. Then
  \begin{align*}
    \Pr[\M(d) \neq j^*] \leq \sum_{j \neq j^*} \frac{2 + \gamma(n_{j^*} - n_j)}{4 \exp(\gamma(n_{j^*} - n_j))}
  \end{align*}
\end{lem}

This allows us to upper bound $q$ for a specific score vector $\mathbf{n}$, and
hence bound specific moments. We take the smaller of the bounds we get
from Theorems~\ref{thm:noisymax} and~\ref{thm:lmgf_data_dep}. We compute these moments
for a few values of $\lambda$ (integers up to
8). Theorem~\ref{thm:property} allows us to add these bounds over
successive steps, and  derive an $(\eps,\delta)$
guarantee from the final $\alpha$. Interested readers
are referred to the script that we used to empirically compute these bounds, which is released
along with our code: {\small\url{https://github.com/tensorflow/models/tree/master/differential_privacy/multiple_teachers}}

Since the privacy moments are themselves now data dependent, the final $\eps$ is
itself data-dependent and should not be revealed. To get around this,
we bound the {\em smooth sensitivity}~\citep{nissim2007smooth} of the
moments and add noise proportional to it to the moments
themselves. This gives us a differentially private estimate of the
privacy cost. Our evaluation in Section~\ref{sec:evaluation} ignores
this overhead and reports the un-noised values of $\eps$. 
Indeed, in our experiments on MNIST and SVHN, the scale of the noise 
one needs to add to the released $\varepsilon$ is smaller 
than 0.5 and 1.0 respectively.

How does the number of teachers affect the privacy cost? Recall that
the student uses a noisy label computed in (\ref{eq:noisy-max}) which
has a parameter $\gamma$. To ensure that the noisy label is likely to
be the correct one, the noise scale $\frac 1 \gamma$ should be small
compared to the the additive gap between the two largest
vales of $n_j$. While the exact dependence of $\gamma$ on the
privacy cost in Theorem~\ref{thm:lmgf_data_dep} is subtle, as a
general principle, a smaller $\gamma$ leads to a smaller privacy
cost. Thus, a larger gap translates to a smaller privacy cost. Since the
gap itself increases with the number of teachers, having more teachers
would lower the privacy cost. This is true up to a point. With $n$
teachers, each teacher only trains on a $\frac 1 n$ fraction of the
training data. For large enough $n$, each teachers will have too
little training data to be accurate.

To conclude, we note that our analysis is rather conservative in that it
pessimistically assumes that, even if just one example in the training set for
one teacher changes, the classifier produced by that teacher may change
arbitrarily. One advantage of our approach, which enables its 
wide applicability, is that our analysis does not require any assumptions about the workings of the teachers.
Nevertheless, we expect that stronger privacy guarantees may perhaps
be established in specific settings---when assumptions can be made on the learning algorithm 
used to train the teachers. 

%% file: evaluation.tex
\section{Evaluation}
\label{sec:evaluation}

In our evaluation of PATE and its generative variant PATE-G, we first train a teacher ensemble for each dataset. The trade-off between the
accuracy and privacy of labels predicted by the ensemble is greatly dependent on
the number of teachers in the ensemble: being able to train a large set of
teachers  is essential to support the injection of noise yielding strong privacy
guarantees while having a limited impact on accuracy. Second, we minimize the
privacy budget spent on learning the student by training it with as few queries
to the  ensemble as possible.

Our experiments use MNIST and the extended SVHN datasets. Our MNIST model stacks two
convolutional layers with max-pooling and one fully connected layer with ReLUs. When trained on the
entire dataset, the non-private model has a $99.18\%$ test accuracy. For SVHN,
we add two hidden layers.\footnote{The model is adapted from
\scriptsize{\url{https://www.tensorflow.org/tutorials/deep_cnn}}} The non-private model
achieves a $92.8\%$ test accuracy, which is shy of the state-of-the-art.
However, we are primarily interested in comparing the private student's accuracy
with the one of a non-private model trained on the entire dataset, for different
privacy guarantees. The source code for reproducing the results in this section is available on GitHub.\footnote{\scriptsize{\url{https://github.com/tensorflow/models/tree/master/differential_privacy/multiple_teachers}}}

\subsection{Training an ensemble of teachers producing private labels}
\label{ssec:eval-teacher}

As mentioned above, compensating the noise introduced by the Laplacian mechanism
presented in Equation~\ref{eq:noisy-max} requires large ensembles. We evaluate
the extent to which the two datasets considered can be partitioned with a
reasonable impact on the performance of individual teachers. Specifically, we
show that for MNIST and SVHN, we are able to train ensembles of $250$ teachers.
Their aggregated predictions are accurate despite the injection of large amounts
of random noise to ensure privacy. The aggregation mechanism output has an
accuracy of $93.18\%$ for MNIST and $87.79\%$ for SVHN, when evaluated on their
respective test sets, while each query has a low privacy budget of
$\varepsilon=0.05$.

\boldpara{Prediction accuracy} All other things being equal,
the number $n$ of teachers is limited by a
trade-off between the classification task's complexity and the available data.
We train $n$ teachers by partitioning the training data $n$-way. 
 Larger values of $n$ lead to larger
  absolute gaps, hence potentially allowing for a larger noise level and stronger privacy guarantees. At
  the same time, a larger $n$ implies a smaller training dataset for
  each teacher, potentially reducing the teacher accuracy. We
empirically find appropriate values of $n$ for the MNIST and SVHN
datasets by measuring the test set 
accuracy of each teacher trained on one of the $n$ partitions of the training
data. We find that even for $n=250$, the average test accuracy of individual
teachers is $83.86\%$ for MNIST and $83.18\%$ for SVHN. The larger size of SVHN
compensates its increased task complexity.

\begin{figure}
	\centering
	\parbox{0.45\textwidth}{
		\includegraphics[width=0.45\textwidth]{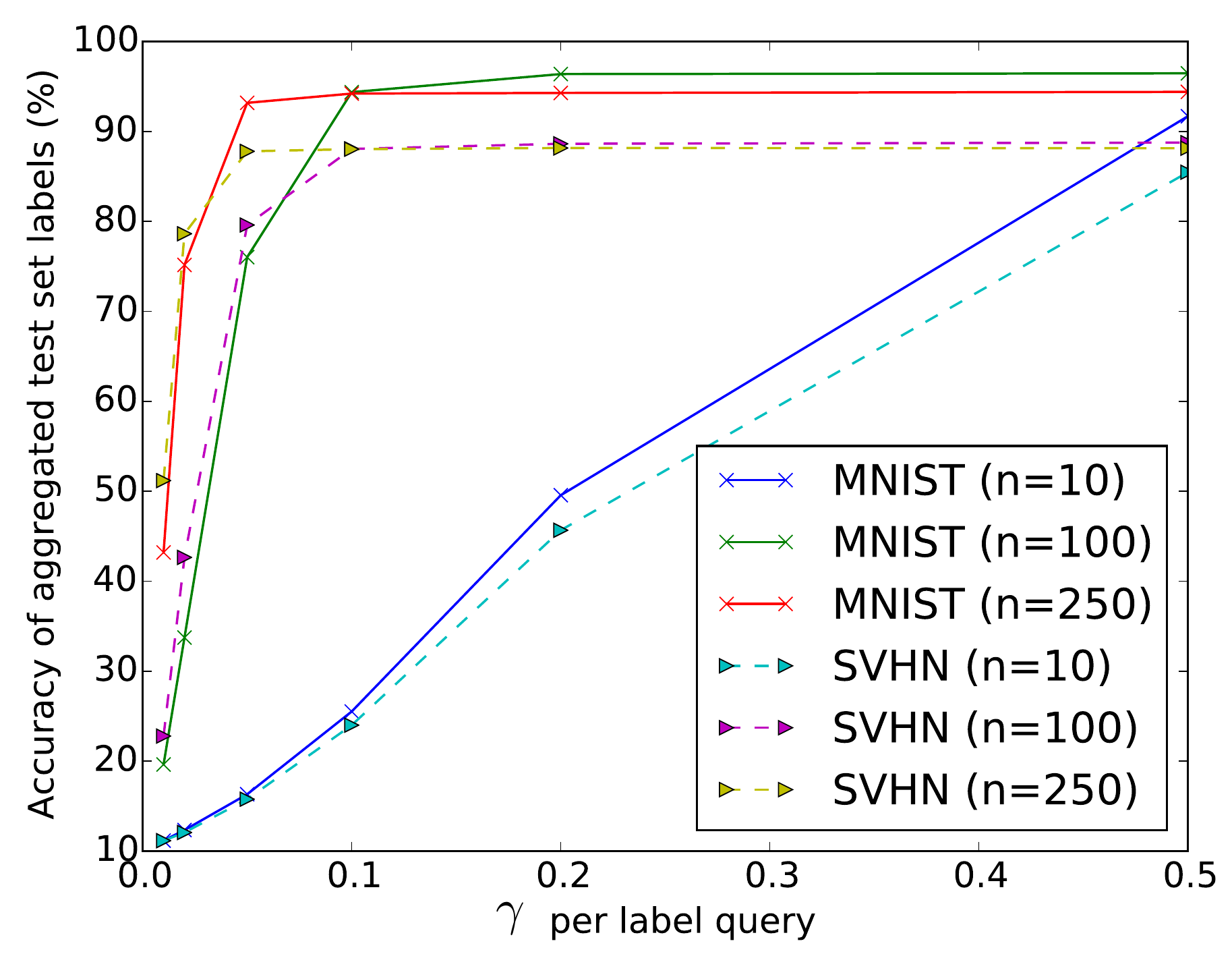}
		\caption{\textbf{How much noise can be injected to a query?}  Accuracy of 
			the noisy aggregation for three MNIST and SVHN teacher ensembles and varying
			$\gamma$ value per query. The noise introduced to achieve a given
			$\gamma$ scales inversely proportionally to the value of $\gamma$:
			small values of $\gamma$ on the left of the axis correspond to large
			noise amplitudes and large $\gamma$ values on the right to small noise.}
		\label{fig:lap-scale-accuracy}
	}
	\qquad
	\begin{minipage}{0.45\textwidth}
		\includegraphics[width=\textwidth]{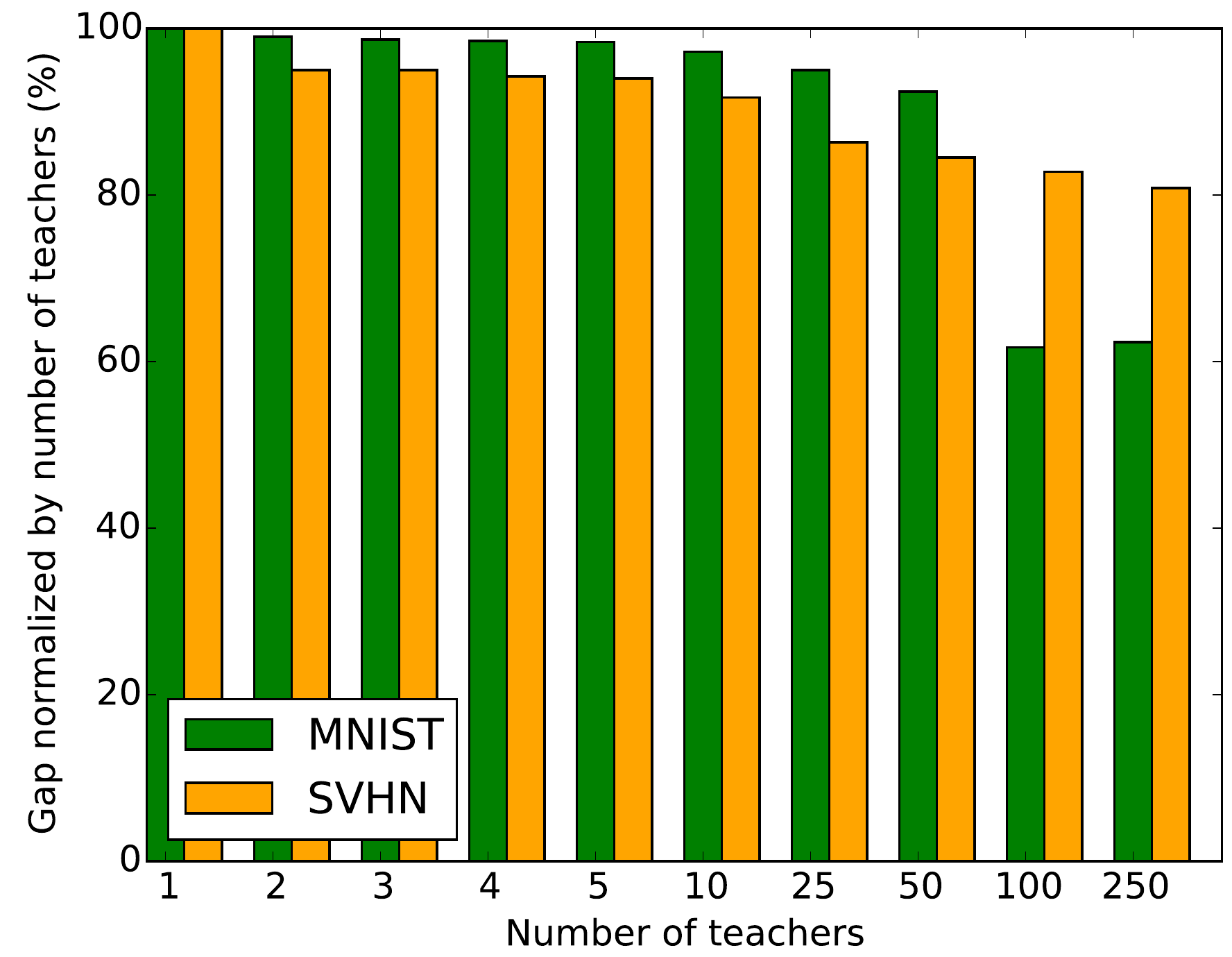}
		\caption{\textbf{How certain is the aggregation of teacher predictions?} Gap
		between the number of votes assigned to the most and second most frequent
		labels normalized by the number of teachers in an ensemble. Larger gaps
		indicate that the ensemble is confident in assigning the labels, and will be
		robust to more noise injection. Gaps were computed by averaging over the test
		data.}
		\label{fig:nb-teachers-gaps}
	\end{minipage}
\end{figure}

\boldpara{Prediction confidence} As outlined in Section~\ref{sec:dp-analysis},
the privacy of predictions made by an ensemble of teachers intuitively requires
that a quorum of teachers generalizing well agree on identical labels. This
observation is reflected by our data-dependent privacy analysis, which provides
stricter privacy bounds when the quorum is strong. We study the disparity of
labels assigned by teachers. In other words, we count
  the number of votes for each possible label, and measure the
  difference in votes between the most popular label and the second
  most popular label, i.e., the \emph{gap}. If the gap is small, introducing noise
during aggregation might change the label assigned from the first to the second.
Figure~\ref{fig:nb-teachers-gaps} shows the gap normalized by the total number
of teachers $n$. As $n$ increases, the gap remains larger than $60\%$
of the teachers, allowing for aggregation mechanisms to output the correct label
in the presence of noise.

\boldpara{Noisy aggregation} For MNIST and SVHN, we consider three ensembles of
teachers with varying number of teachers $n\in\{10,100,250\}$. For each of them,
we perturb the vote counts with Laplacian noise of inversed scale $\gamma$
ranging between $0.01$ and $1$. This choice is justified below in
Section~\ref{ssec:eval-student}. We report in
Figure~\ref{fig:lap-scale-accuracy} the accuracy of test set labels inferred by
the noisy aggregation mechanism for these values of $\varepsilon$. Notice that
the number of teachers needs to be large to compensate for the impact of noise
injection on the accuracy.

\subsection{Semi-supervised training of the student with privacy}
\label{ssec:eval-student}

The noisy aggregation mechanism labels the student's unlabeled training set in a
privacy-preserving fashion. To reduce the privacy budget spent on student
training, we are interested in making as few label queries to the teachers as
possible. We therefore use the semi-supervised training approach described
previously. Our MNIST and SVHN students with $(\varepsilon,\delta)$ differential
privacy of $(2.04, 10^{-5})$ and $(8.19, 10^{-6})$ achieve accuracies of
$98.00\%$ and $90.66\%$. These results improve the differential privacy
state-of-the-art for these datasets. \citet{abadi2016deep} previously obtained
$97\%$ accuracy with a $(8, 10^{-5})$ bound on MNIST, starting from an inferior baseline model without privacy. \citet{shokri2015privacy}
reported about $92\%$ accuracy on SVHN with $\varepsilon>2$ per model parameter
and a model with over $300\mbox{,}000$ parameters. Naively, this corresponds to
a total $\varepsilon>600\mbox{,}000$.

We apply semi-supervised learning with GANs to our problem using the following
setup for each dataset. In the case of MNIST, the student has access to
$9\mbox{,}000$ samples, among which a subset of either $100$, $500$, or
$1\mbox{,}000$ samples are labeled using the noisy aggregation mechanism
discussed in Section~\ref{ssec:teacher}. Its performance is evaluated on the
$1\mbox{,}000$ remaining samples of the test set. Note that this may increase
the variance of our test set accuracy measurements, when compared to those computed over
the entire test data. For the MNIST
dataset, we randomly shuffle the test set to ensure that the different classes
are balanced when selecting the (small) subset labeled to train the student. For
SVHN, the student has access to $10\mbox{,}000$ training inputs, among which it
labels $500$ or $1\mbox{,}000$ samples using the noisy aggregation mechanism.
Its performance is evaluated on the remaining $16\mbox{,}032$ samples. For both
datasets, the ensemble is made up of $250$ teachers. We use Laplacian scale of
$20$ to guarantee an individual query privacy bound of $\varepsilon=0.05$. These
parameter choices are motivated by the results from
Section~\ref{ssec:eval-teacher}. 

In Figure~\ref{fig:gans}, we report the values
of the $(\varepsilon, \delta)$ differential privacy guarantees provided and the
corresponding student accuracy, as well as the number of queries made by each
student. The MNIST student is able to learn a $98\%$ accurate model, which is
shy of $1\%$ when compared to the accuracy of a model learned with the entire
training set,  with only $100$ label queries. This results in a strict
differentially private bound of $\varepsilon=2.04$ for a failure probability
fixed at $10^{-5}$. The SVHN student achieves $90.66\%$ accuracy, which is also
comparable to the $92.80\%$ accuracy of one teacher learned with the entire
training set. The corresponding privacy bound is $\varepsilon=8.19$, which is
higher than for the MNIST dataset, likely because of the larger number of
queries made to the aggregation mechanism.

We observe that our private student outperforms the aggregation's output in terms
of accuracy, with or without the injection of Laplacian noise. While this shows
the power of semi-supervised learning, the student may not learn as well on different
kinds of data (e.g., medical data), where categories are not explicitly designed by humans to be
salient in the input space.
Encouragingly,
as Appendix~\ref{ap:uci} illustrates,
the PATE approach can be successfully applied to at least some examples of such data.

\begin{figure}
	\centering
	\begin{tabular}{c|c|c|c|c|c}
		\textbf{Dataset} & \textbf{$\varepsilon$}  & \textbf{$\delta$} & 
		\textbf{Queries} & \textbf{Non-Private Baseline} & \textbf{Student Accuracy}  \\ \hline \hline 
		MNIST & 2.04 & $10^{-5}$\rule[.05ex]{0em}{1em} & 100 & 99.18\%  & 98.00\%   \\ \hline
		MNIST & 8.03 & $10^{-5}$\rule[.05ex]{0em}{1em}  & 1000 & 99.18\%  & 98.10\%   \\ \hline \hline 
		SVHN & 5.04 & $10^{-6}$\rule[.05ex]{0em}{1em}  & 500 &  92.80\%  & 82.72\%   \\ \hline 
		SVHN & 8.19 & $10^{-6}$\rule[.05ex]{0em}{1em}  & 1000 &  92.80\%  & 90.66\%   \\ \hline 
	\end{tabular}
	\caption{\textbf{Utility and privacy of the semi-supervised students:} each row 
		is a variant of the student model trained with generative adversarial networks
		in a semi-supervised way, with a different number of label queries made to the
		teachers through the noisy aggregation mechanism. The last column reports the
		accuracy of the student and the second and third column the bound
		$\varepsilon$ and failure probability $\delta$ of the $(\varepsilon, \delta)$
		differential privacy guarantee.
	}
	\label{fig:gans}
\end{figure}

%% file: related-work.tex
\section{Discussion and related work}
\label{sec:related-work}

Several privacy definitions are found in the literature. For instance,
\emph{k-anonymity} requires information about an individual to be
indistinguishable from at least $k-1$ other individuals in the dataset~\citep{sweeney2002k}.
However, its lack of randomization gives rise to
caveats~\citep{dwork2014algorithmic}, and attackers can infer properties of the
dataset~\citep{aggarwal2005k}. An alternative definition, \emph{differential
privacy}, established itself as a rigorous standard for providing privacy
guarantees~\citep{dwork2006calibrating}. In contrast to $k$-anonymity,
differential privacy is a property of the randomized algorithm and not the
dataset itself.

A variety of approaches and mechanisms can guarantee differential privacy.
\citet{erlingsson2014rappor} showed that randomized response, introduced
by~\citet{warner1965randomized}, can protect crowd-sourced data collected from
software users to compute statistics about user behaviors. Attempts to provide
differential privacy for machine learning models led to a series of efforts on
shallow machine learning models, including work
by~\citet{bassily2014differentially,chaudhuri2009privacy,pathak2011privacy,song2013stochastic}, and~\citet{wainwright2012privacy}.

A privacy-preserving distributed SGD algorithm was introduced
by~\citet{shokri2015privacy}. It applies to non-convex models. However, its
privacy bounds are given per-parameter, and the large number of parameters
prevents the technique from providing a meaningful privacy guarantee.
\citet{abadi2016deep} provided stricter bounds on the privacy loss induced by a
noisy SGD by introducing the moments accountant. In comparison with these
efforts, our work increases the accuracy of a private MNIST model from $97\%$ to
$98\%$ while improving the privacy bound $\eps$ from $8$ to $1.9$.
Furthermore, the PATE approach is independent of the learning algorithm, unlike this
previous work. Support
for a wide range of architecture and training algorithms allows us to obtain
good privacy bounds on an accurate and private SVHN model. However, this comes
at the cost of assuming that non-private unlabeled data is available, an
assumption that is not shared by~\citep{abadi2016deep,shokri2015privacy}.

\citet{pathak2010multiparty} first discussed secure multi-party aggregation of
locally trained classifiers for a global classifier hosted by a trusted
third-party. \citet{hamm2016learning} proposed the use of knowledge transfer
between a collection of models trained on individual devices into a single model
guaranteeing differential privacy. Their work studied linear student models with convex and
continuously differentiable losses, bounded and $c$-Lipschitz derivatives, and
bounded features. The PATE approach of this paper is not constrained to such
applications, but is more generally applicable.

Previous work also studied
semi-supervised knowledge transfer from private models. For instance, \citet{jagannathan2013semi} learned privacy-preserving
random forests.
A key difference is that their approach is tailored to decision trees.
PATE works well for the specific case of decision trees, 
as demonstrated in Appendix~\ref{ap:uci},
and is also applicable to other machine learning algorithms, including more complex ones.
Another key difference is that \citet{jagannathan2013semi}
modified the classic model of a decision tree to include the Laplacian mechanism.
Thus, the privacy guarantee does not come from the disjoint sets of training
data analyzed by different decision trees in the random forest, but rather from
the modified architecture. In contrast, partitioning is essential to the
privacy guarantees of the PATE approach.

%% file: conclusions.tex
\section{Conclusions}

To protect the privacy of sensitive training data, this paper has advanced a
learning strategy and a corresponding privacy analysis. The PATE approach is based on knowledge
aggregation and transfer from ``teacher'' models, trained on disjoint data, to a
``student'' model whose attributes may be made public. In combination, the
paper's techniques demonstrably achieve excellent utility on the MNIST and SVHN
benchmark tasks, while simultaneously providing a formal, state-of-the-art bound
on users' privacy loss. While our results are not without limits---e.g., they
require disjoint training data for a large number of teachers (whose number is likely to increase for tasks with many output classes)---they are
encouraging, and highlight the advantages of combining semi-supervised learning
with precise, data-dependent privacy analysis, which will hopefully trigger
further work. In particular, such future work may further investigate whether or not our
semi-supervised approach will also reduce teacher queries for tasks other than
MNIST and SVHN, for example when the discrete output categories are not as
distinctly defined by the salient input space features.

A key advantage is that this paper's techniques establish a precise guarantee of
training data privacy in a manner that is both intuitive and rigorous.
Therefore, they can be appealing, and easily explained, to both an expert and
non-expert audience. However, perhaps equally compelling are the techniques'
wide applicability. Both our learning approach and our analysis
methods are ``black-box,'' i.e., independent of the learning algorithm for
either teachers or students, and therefore apply, in general, to 
non-convex, deep learning, and other learning methods. Also, because our
techniques do not constrain the selection or partitioning of training data, they
apply when training data is naturally and non-randomly partitioned---e.g.,
because of privacy, regulatory, or competitive concerns---or when each teacher
is trained in isolation, with a different method. We look forward to
such further applications, for example on RNNs and other sequence-based models.

%% file: ap-privacy-analysis-short.tex
\section{Missing details on the analysis}
\label{ap:privacy-analysis}

We provide missing proofs from Section~\ref{sec:dp-analysis}.
\newtheorem*{thm:lmgfdatadep}{Theorem \ref{thm:lmgf_data_dep}}
\begin{thm:lmgfdatadep}
	Let $\M$ be $(2\gamma, 0)$-differentially private and $q \geq \Pr[\M(d) \neq \outcome^*]$ for some outcome $\outcome^*$. Let $l,\gamma \geq 0$ and $q <\frac{e^{2\gamma}-1}{e^{4\gamma}-1}$. Then for any $\textsf{aux}$ and any neighbor $d'$ of $d$, $\M$ satisfies
	\begin{align*}
	\alpha(l; \textsf{aux}, d, d') \leq \log ((1-q)\Big(\frac{1-q}{1-e^{2\gamma}q}\Big)^l + q\exp(2\gamma l)).
	\end{align*}
\end{thm:lmgfdatadep}

\begin{proof}
	Since $\M$ is $2\gamma$-differentially private, for every outcome $o$, $\frac{Pr[M(d)=\outcome]}{Pr[M(d')=\outcome]} \leq \exp(2\gamma)$. Let $q' = Pr[M(d) \neq \outcome^*]$. Then $Pr[M(d') \neq \outcome^*] \leq \exp(2\gamma) q'$. Thus
	\begin{align*}
	\exp(\alpha(l; \textsf{aux},d,d')) &= \sum_{\outcome} \Pr[M(d)=\outcome] \Big(\frac{\Pr[M(d)=\outcome]}{\Pr[M(d')=\outcome]}\Big)^l\\
	&= \Pr[M(d)=\outcome^*] \Big(\frac{\Pr[M(d)=\outcome^*]}{\Pr[M(d')=\outcome^*]}\Big)^l + \sum_{\outcome \neq \outcome^*} \Pr[M(d)=\outcome] \Big(\frac{\Pr[M(d)=\outcome]}{\Pr[M(d')=\outcome]}\Big)^l\\
	&\leq (1-q')\Big(\frac{1-q'}{1-e^{2\gamma}q'}\Big)^l + \sum_{\outcome \neq \outcome^*} \Pr[M(d)=\outcome] (e^{2\gamma})^l\\
	&\leq  (1-q')\big(\frac{1-q'}{1-e^{2\gamma}q'}\Big)^l + q'e^{2\gamma l}.
	\end{align*}
        Now consider the function
        \begin{align*}
          f(z) &= (1-z)\Big(\frac{1-z}{1-e^{2\gamma}z}\Big)^l + z e^{2\gamma l}.
        \end{align*}
        We next argue that this function is non-decreasing in $(0,\frac{e^{2\gamma}-1}{e^{4\gamma}-1})$ under the conditions of the lemma. Towards this goal, define
        \begin{align*}
          g(z,w) &= (1-z)\Big(\frac{1-w}{1-e^{2\gamma}w}\Big)^l + z e^{2\gamma l},
        \end{align*}
        and observe that $f(z) = g(z,z)$. We
        can easily verify by differentiation that $g(z,w)$ is
        increasing individually in $z$ and in $w$ in the range of
        interest. This implies that $f(q') \leq f(q)$ completing the proof.
\end{proof}
\newtheorem*{lem:boundonq}{Lemma \ref{lem:bound_on_q}}
\begin{lem:boundonq}
  Let $\mathbf{n}$ be the label score vector for a database $d$ with $n_{j^*} \geq n_j$ for all $j$. Then
  \begin{align*}
    \Pr[\M(d) \neq j^*] \leq \sum_{j \neq j^*} \frac{2 + \gamma(n_{j^*} - n_j)}{4 \exp(\gamma(n_{j^*} - n_j))}
  \end{align*}
\end{lem:boundonq}
\begin{proof}
The probability that $n_{j^*} + Lap(\frac 1 \gamma) < n_j + Lap(\frac 1 \gamma)$
is equal to the probability that the sum of two independent $Lap(1)$ random
variables exceeds $\gamma(n_{j^*} - n_j)$. The sum of two independent $Lap(1)$
variables has the same distribution as the difference of two $Gamma(2, 1)$
random variables. Recalling that the $Gamma(2,1)$ distribution has pdf
$xe^{-x}$, we can compute the pdf of the difference via convolution as
\begin{align*}
\int_{y=0}^\infty (y+|x|)e^{-y-|x|} y e^{-y}~dy = \frac{1}{e^{|x|}}\int_{y=0}^\infty (y^2+y|x|) e^{-2y}~dy = \frac{1+|x|}{4e^{|x|}}.
\end{align*} 
The probability mass in the tail can then be computed by integration as $\frac{2 + \gamma(n_{j^*} - n_j)}{4 \exp(\gamma(n_{j^*} - n_j)}$. Taking a union bound over the various candidate $j$'s gives the claimed bound.
\end{proof}

%% file: ap-learning-student.tex
\section{Appendix: Training the student with minimal teacher queries}
\label{ap:student-learning}

In this appendix, we describe approaches that were considered to reduce the
number of queries made to the teacher ensemble by the student during its
training. As pointed out in Sections~\ref{sec:dp-analysis}
and~\ref{sec:evaluation}, this effort is motivated by the direct impact of
querying on the total privacy cost associated with student training. The first
approach is based on \emph{distillation}, a technique used for knowledge
transfer and model compression~\citep{hinton2015distilling}. The three other
techniques considered were proposed in the context of \emph{active learning},
with the intent of identifying training examples most useful for learning. 
In
Sections~\ref{sec:approach} and~\ref{sec:evaluation}, we described
semi-supervised learning, which yielded the best results. 
The student models in this appendix differ from those in Sections~\ref{sec:approach} and~\ref{sec:evaluation},
which were trained using GANs.
In contrast, all
students in this appendix
were learned in a fully supervised fashion from a
subset of public, labeled examples. Thus, the learning goal was
to identify the subset of labels yielding the best learning performance. 

\subsection{Training Students using Distillation}

Distillation is a knowledge transfer technique introduced as a means of compressing
large models into smaller ones, while retaining their
accuracy~\citep{bucilua2006model,hinton2015distilling}. This is for instance
useful to train models in data centers before deploying compressed variants in
phones. The transfer is accomplished by training the smaller model on data that
is labeled with probability vectors produced by the first model, which encode
the knowledge extracted from training data. Distillation is parameterized by a
\emph{temperature} parameter $T$, which controls the smoothness of probabilities
output by the larger model: when produced at small temperatures, the vectors are
discrete, whereas at high temperature, all classes are assigned non-negligible
values. Distillation is a natural candidate to compress the knowledge acquired
by the ensemble of teachers, acting as the large model, into a student, which is
much smaller with $n$ times less trainable parameters compared to the $n$
teachers.

\begin{figure}[t]
	\centering
	\includegraphics[width=0.8\textwidth]{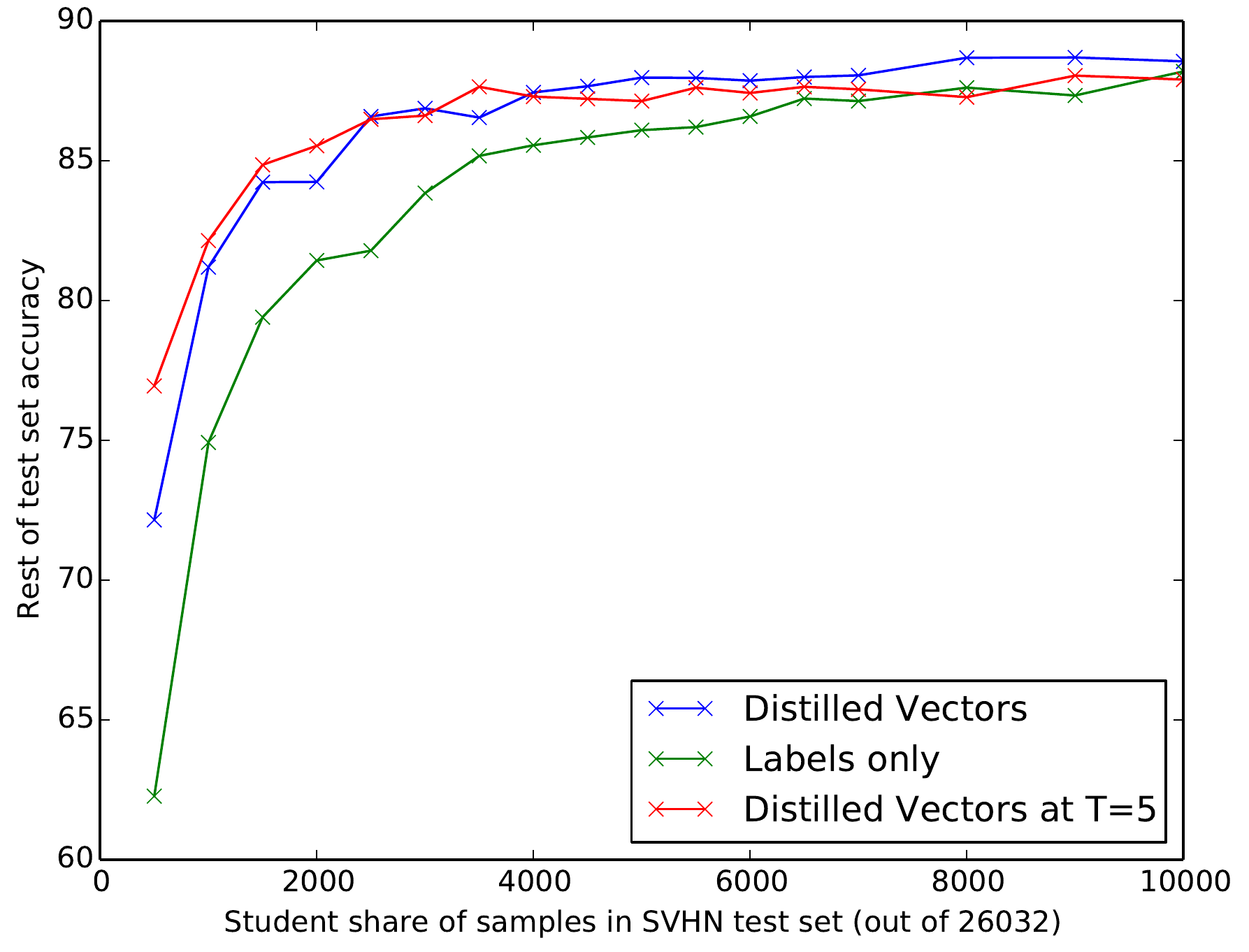}
	\caption{Influence of distillation on the accuracy of the SVHN student 
		trained with respect to the initial number of training samples available to 
		the student. The student is learning from $n=50$ teachers, whose predictions 
		are aggregated without noise: in case where only the label is returned, we 
		use plurality, and in case a probability vector is returned, we sum the 
		probability vectors output by each teacher before normalizing the resulting 
		vector.}
	\label{fig:svhn-student-distillation}
\end{figure}

To evaluate the applicability of distillation, we consider the ensemble of
$n=50$ teachers for SVHN. In this experiment, we do not add noise to the vote
counts when aggregating the teacher predictions. We compare the accuracy of
three student models: the first is a baseline trained with labels obtained by
plurality, the second and third are trained with distillation at $T\in\{1,5\}$.
We use the first $10\mbox{,}000$ samples from the test set as unlabeled data.
Figure~\ref{fig:svhn-student-distillation} reports the accuracy of the student
model on the last $16\mbox{,}032$ samples from the test set, which were not
accessible to the model during training. It is plotted with respect to the
number of samples used to train the student (and hence the number of queries
made to the teacher ensemble). Although applying distillation yields classifiers
that perform more accurately, the increase in accuracy is too limited to justify
the increased privacy cost of revealing the entire probability vector output by
the ensemble instead of simply the class assigned the largest number of votes.
Thus, we turn to an investigation of active learning.

\subsection{Active Learning of the Student}

Active learning is a class of techniques that aims to identify and prioritize
points in the student's training set that have a high potential to contribute to
learning~\citep{angluin1988queries,baum1991neural}. If the label of an input in
the student's training set can be predicted confidently from what we have
learned so far by querying the teachers, it is intuitive that querying it is not
worth the privacy budget spent. In our experiments, we made several attempts
before converging to a simpler final formulation.

\boldpara{Siamese networks} Our first attempt was to train a pair of siamese
networks, introduced by~\citet{bromley1993signature} in the context of one-shot
learning and later improved by~\citet{kochsiamese}. The siamese networks take
two images as input and return $1$ if the images are equal and $0$ otherwise.
They are two identical networks trained with shared parameters to force them to
produce similar representations of the inputs, which are then compared using a
distance metric to determine if the images are identical or not. Once the
siamese models are trained, we feed them a pair of images where the first
is unlabeled and the second labeled. If the unlabeled image is confidently
matched with a known labeled image, we can infer the class of the unknown image
from the labeled image. In our experiments, the siamese
networks were able to say whether two images are identical or not, but did not
generalize well: two images of the same class did not receive sufficiently
confident matches. We also tried a variant of this approach where we trained the
siamese networks to output $1$ when the two images are of the same class and $0$
otherwise, but the learning task proved too complicated to be an effective means
for reducing the number of queries made to teachers.

\boldpara{Collection of binary experts} Our second attempt was to train a collection of binary experts, one per class. An expert for class $j$ is trained to output $1$ if the sample is in class $j$ and $0$ otherwise. We first trained the binary experts by making an initial batch of
queries to the teachers. Using the experts, we then selected available unlabeled
student training points that had a candidate label score below $0.9$ and at
least $4$ other experts assigning a score above $0.1$. This gave us about $500$
unconfident points for $1700$ initial label queries. After labeling these
unconfident points using the ensemble of teachers, we trained the student. 
Using binary experts
improved the student's accuracy when compared to the student trained on
arbitrary data with the same number of teacher queries. The absolute increases
in accuracy were however too limited---between $1.5\%$ and $2.5\%$.

\boldpara{Identifying unconfident points using the student} This last attempt
was the simplest yet the most effective. Instead of using binary experts to
identify student training points that should be labeled by the teachers, we used
the student itself. We asked the student to make predictions on each unlabeled
training point available. We then sorted these samples by increasing values of
the maximum probability assigned to a class for each sample. We queried the
teachers to label these unconfident inputs first and trained the student again
on this larger labeled training set. This improved the accuracy of the student
when compared to the student trained on arbitrary data. 
For the same number of teacher queries, the absolute increases in accuracy
of the student trained on unconfident inputs first
 when compared to the student trained on arbitrary data were in the order of
$4\%-10\%$.

%% file: ap-uci.tex
\section{Appendix: Additional experiments on the UCI Adult and Diabetes datasets}
\label{ap:uci}

In order to further demonstrate the general applicability of our approach, we
performed experiments on two additional datasets. While our experiments on MNIST
and SVHN in Section~\ref{sec:evaluation} used convolutional neural networks and
GANs, here we use random forests to train our teacher and student models for
both of the datasets. Our new results on these datasets show that, despite
the differing data types and architectures, we are able to provide meaningful
privacy guarantees.

\boldpara{UCI Adult dataset} The UCI Adult dataset is made up of census data,
and the task is to predict when individuals make over \$50k per year. Each input
consists of 13 features (which include the age, workplace, education,
occupation---see the UCI website for a full
list\footnote{\scriptsize{\url{https://archive.ics.uci.edu/ml/datasets/Adult}}}). The only
pre-processing we apply to these features is to map all categorical features to
numerical values by assigning an integer value to each possible category. 
The model is a random forest provided by the \texttt{scikit-learn} Python
package. When training both our teachers and student, we keep all the default
parameter values, except for the number of estimators, which we set to $100$.
The data is split between a training set of $32\mbox{,}562$ examples, and a test set of
$16\mbox{,}282$ inputs.

\boldpara{UCI Diabetes dataset} The UCI Diabetes dataset includes de-identified
records of diabetic patients and corresponding hospital outcomes, which we use
to predict whether diabetic patients were readmitted less than 30 days after
their hospital release. To the best of our knowledge, no particular
classification task is considered to be a standard benchmark for this dataset.
Even so, it is valuable to consider whether our approach is
applicable to the likely classification tasks, such as readmission,
since this dataset is collected in a medical environment---a setting where privacy
concerns arise frequently. We select a subset of $18$ input features from the
$55$ available in the dataset (to avoid features with missing values) and form a dataset
balanced between the two output classes (see the UCI website for more
details\footnote{\scriptsize{\url{https://archive.ics.uci.edu/ml/datasets/Diabetes+130-US+hospitals+for+years+1999-2008}}}). 
In class $0$, we include all patients that were readmitted in a 30-day window,
while class $1$ includes all patients that were readmitted after 30 days or
never readmitted at all. Our balanced dataset contains $34\mbox{,}104$
training samples and $12\mbox{,}702$ evaluation samples. We use a random forest model identical to
the one described above in the presentation of the Adult dataset.

\boldpara{Experimental results} We apply our approach described in Section~\ref{sec:approach}.
For both datasets, we train ensembles of $n=250$ random
forests on partitions of the training data. We then use the noisy aggregation
mechanism, where vote counts are perturbed with Laplacian noise of scale $0.05$
to privately label the first $500$ test set inputs. We train the student random
forest on these $500$ test set inputs and evaluate it on the last $11\mbox{,}282$ test
set inputs for the Adult dataset, and $6\mbox{,}352$ test set inputs for the Diabetes dataset. These
numbers deliberately leave out some of the test set, which allowed us to observe
how the student performance-privacy trade-off was impacted by varying the number
of private labels, as well as the Laplacian scale used when computing these labels.

For the Adult dataset,
we find that our student model achieves an $83\%$ accuracy for an $(\varepsilon,
\delta) = (2.66, 10^{-5})$ differential privacy bound. Our non-private model on
the dataset achieves $85\%$ accuracy, which is comparable to the
state-of-the-art accuracy of $86\%$ on this dataset~\citep{poulos2016missing}.
For the Diabetes dataset, we find that our privacy-preserving student model achieves a $93.94\%$
accuracy for a $(\varepsilon, \delta) = (1.44, 10^{-5})$ differential privacy
bound.  Our non-private model on the dataset achieves $93.81\%$ accuracy.